\documentclass[journal]{IEEEtran}
\usepackage[noadjust]{cite}
\usepackage{amsmath}
\usepackage{color}
\usepackage{epsfig}
\usepackage{enumitem}
\usepackage{graphicx}
\usepackage{subcaption}
\usepackage[flushleft]{threeparttable}
\usepackage[hyphenbreaks]{breakurl}
\usepackage[hyphens]{url}
\graphicspath{{Figures/}}

\begin{document}
\bstctlcite{IEEEexample:BSTcontrol}
\title{A Fusion Framework for Camouflaged Moving Foreground Detection in the Wavelet Domain}

\author{Shuai~Li,
        Dinei~Florencio,~\IEEEmembership{Fellow,~IEEE,}
        Wanqing~Li,~\IEEEmembership{Senior Member,~IEEE,}
        Yaqin~Zhao,
        and~Chris~Cook
\thanks{Manuscript received June 23, 2017; revised Nov. 3, 2017; accepted Apr. 10, 2018. This work was partially supported by the Global Challenge Project 2015-2016, Assistive Systems for the Ageing, of University of Wollongong.}%
\thanks{S. Li, W. Li and C. Cook are with University of Wollongong, NSW 2522, Australia. Email: \{sl669, wanqing, ccook\}@uow.edu.au.}
\thanks{D. Florencio is with Microsoft Research, WA 98052, USA. Email: dinei@microsoft.com.}
\thanks{Y. Zhao is with Nanjing Forestry University, Nanjing 210037, Jiangsu Province, China. Email: yaqinzhao@163.com.}}%

\markboth{IEEE Transactions on Image Processing}%
{Li \MakeLowercase{\textit{et al.}}: A Fusion Framework for Camouflaged Moving Foreground Detection in Wavelet Domain}

\maketitle

\begin{abstract}
Detecting camouflaged moving foreground objects has been known to be difficult due to the similarity between the foreground objects and the background. Conventional methods cannot distinguish the foreground from background due to the small differences between them and thus suffer from under-detection of the camouflaged foreground objects. In this paper, we present a fusion framework to address this problem in the wavelet domain. We first show that the small differences in the image domain can be highlighted in certain wavelet bands. Then the likelihood of each wavelet coefficient being foreground is estimated by formulating foreground and background models for each wavelet band. The proposed framework effectively aggregates the likelihoods from different wavelet bands based on the characteristics of the wavelet transform. Experimental results demonstrated that the proposed method significantly outperformed existing methods in detecting camouflaged foreground objects. Specifically, the average F-measure for the proposed algorithm was 0.87, compared to 0.71 to 0.8 for the other state-of-the-art methods.
\end{abstract}



\section{Introduction}
\IEEEPARstart{I}{n} many video processing and computer vision systems, moving foreground object detection is an important pre-processing step for further in-depth analysis. Especially in video-surveillance applications, it is often used to detect persons and vehicles before performing intrusion detection, tracking, etc. A commonly used approach for foreground detection is background subtraction, where a background model is first formulated and then the foreground object is detected by comparing the current frame and the background model. 

There are many background subtraction methods \cite{piccardi2004background, radke2005image, bouwmans2008background, bouwmans2009subspace, cristani2010background, bouwmans2011recent, bouwmans2014robust, sobral2014comprehensive, bouwmans2014traditional, jeeva2015survey, xu2016background, shahbaz2017recent, bouwmans2017decomposition} in the literature. Most of them are developed under the assumption that foreground and background show visually distinct characteristics and thus the foreground can be detected once a good background model is obtained. However, there are cases of camouflage in color \cite{bouwmans2016role} where the foreground share similar color as the background. For example, in Fig. \ref{CamExample}, a person wears clothes that have similar colors as the background wall. In such a case, the intensity difference between the foreground and background are very small and the foreground cannot be easily detected by the existing methods. This paper addresses the problem of ``camouflage in color'' and it is simply referred to as ``camouflage'' in the rest of the paper.

In this paper, we propose to address the camouflaged moving foreground detection problem in the wavelet domain. It is first shown that visually small/unnoticeable differences in the image domain may become larger and detectable in certain wavelet bands. Based on this, a framework is proposed to fuse the likelihood of being foreground in the multiple wavelet bands. Following the properties of the wavelet transform, different fusion methods are developed for different levels and bands. The proposed fusion framework is denoted by ``FWFC'' (\textbf{F}usion in the \textbf{W}avelet domain for \textbf{F}oreground detection in \textbf{C}amouflaged scenes). Experimental results have shown that FWFC can perform significantly better than the existing methods in the camouflage scenes.

Part of this work has appeared in the conference paper \cite{Li2017Foreground}, where a texture guided weighted voting scheme was proposed. However, significant extension has been made in this paper. Specifically, the new contributions in this paper can be summarized as follows.

\begin{itemize}
\item Unlike~\cite{Li2017Foreground} where the background model is formulated in the image domain and wavelet transform is applied to the difference between the estimated background and the current frame, we have directly formulated the background model in the wavelet domain, which provides a more robust difference between the current frame and the background model in each wavelet band.
\item Instead of making decisions in all bands and fusing the decisions as in \cite{Li2017Foreground}, the proposed method formulates the likelihood of being foreground in each wavelet band and fuses them in a rigorous way by considering the characteristics of the wavelet.
\item A larger dataset containing 10 videos of camouflaged scenes is collected with manually labelled groundtruth for all frames.  Experiments on more datasets and comparisons with more existing methods compared to~\cite{Li2017Foreground} were conducted to verify the efficacy of the proposed method.
\end{itemize} 

The rest of the paper is organized as follows. In Section \ref{relatedwork}, the related work is reviewed. The motivation and overview of the proposed approach is described in Section \ref{motandove} and the details of the proposed approach are explained in Section \ref{propmethod}. Experiments are shown in Section \ref{experimentsec}, followed by the conclusion in Section \ref{concsec}.

\section{Related work}
\label{relatedwork}
Over the past decades, many methods have been reported for background subtraction, including statistical models, fuzzy models, neural models, subspace models, low rank models, sparse models and transform domain models. Here we only review the most relevant works including the statistical background models especially the Gaussian mixture models (GMM), the transform domain methods especially those employing the wavelet transform, and the existing background models specifically designed for camouflaged scenes. Other approaches, which go beyond the scope of this paper, can be found in several survey papers \cite{piccardi2004background, radke2005image, bouwmans2008background, bouwmans2009subspace, cristani2010background, bouwmans2011recent, bouwmans2014robust, sobral2014comprehensive, bouwmans2014traditional, jeeva2015survey, xu2016background, shahbaz2017recent, bouwmans2017decomposition}. 

\subsection{Statistical Background Models}
One of the most popular background modelling methods is the Gaussian mixture models (GMM) \cite{stauffer1999adaptive, hayman2003statistical}. It models the intensity distribution of each pixel using a summation of weighted Gaussian distributions. Specifically, a predefined number of Gaussian components are used \cite{kaewtrakulpong2002improved}, and the Gaussian components are ordered based on the weight and the variance. The first $B$ components with total probabilities over a defined threshold are used as the current background model. When a new pixel is processed, it is considered to be a foreground pixel if its value cannot be well described by the background Gaussian distributions. These distributions are then updated with the current pixel using an online expectation-minimization algorithm. It is also known as MOG in openCV \cite{opencv_library}. In \cite{zivkovic2006efficient}, an improved adaptive Gaussian Mixture Model was proposed which can adaptively estimate the number of the Gaussian components needed, known as MOG2 in openCV. In \cite{tuzel2005bayesian}, 3D multivariate Gaussian distributions are used to formulate a multiple layers model and the mean and variance of the distributions are obtained based on a recursive Bayesian learning approach. 

In addition to the above statistical models based on parametric distributions, there are also methods using non-parametric models such as kernel density estimation \cite{elgammal2000non, elgammal2002background, han2008sequential, Bouwmans2010Statistical, narayana2012background, cuevas2017detection, berjon2018real}. The kernel density estimation (KDE) based methods \cite{elgammal2002background} estimate the probability density function for each pixel from many recent samples ($N$) over time using a kernel function. These methods can be memory consuming considering that $N$ frames have to be kept in memory. Many improvements have been made on the KDE methods, including changing the kernel function. In \cite{han2008sequential}, a sequential kernel density approximation method was proposed based on the mean-shift mode finding algorithm and the density modes are sequentially propagated over time. In \cite{berjon2018real}, the foreground is also modelled and further tracked over time based on a particle filter. A selective analysis strategy was further proposed to reduce the computation. Other than the above KDE based methods, there are also non-parametric methods \cite{wang2007consensus, barnich2011vibe, hofmann2012background} using a history of recently observed pixel values for background modelling instead of using the distribution of the pixels. The foreground decision is made based on the number of the pixels in the background set that the current pixel is close to. Different background update rules have been proposed in the literature. In SACON (sample consensus) \cite{wang2007consensus}, the history of background image is updated using a first-in first-out strategy, while in ViBe (visual background extractor) \cite{barnich2011vibe} and PBAS (pixel-based adaptive segmenter) \cite{hofmann2012background}, the history is updated by a random scheme based on fixed or adaptively changed randomness parameters. 

Another category of statistical models employ subspace learning methods \cite{oliver2000bayesian, de2003framework}. In \cite{oliver2000bayesian}, subspace learning using principal component analysis (SL-PCA) is applied on $N$ images, and the background model is represented by the mean image and the projection matrix comprising the first $p$ significant eigenvectors of PCA. The foreground is obtained by computing the difference between the input image and its reconstruction. Recently, lots of methods have been proposed to formulate the background model via a robust subspace using a low-rank and sparse decomposition such as robust principal component analysis \cite{candes2011robust}. The background is modelled by the low-rank subspace that gradually changes over time while the foreground consists of the correlated sparse outliers. In order to consider the spatial connection of the foreground sparse pixel, in \cite{liu2015background}, a class of structured sparsity-inducing norms were introduced to model the moving objects in videos. In \cite{zhou2013moving}, a framework named detecting contiguous outliers in the low-rank representation (DECOLOR) was proposed where the object detection and background learning are integrated into a single optimization process solved by an alternating algorithm. In \cite{shakeri2016corola}, an online sequential framework named contiguous outliers representation via online low-rank approximation (COROLA) was proposed to detect moving objects and learn the background model at the same time. It works iteratively on each image of the video to extract foreground objects by exploiting the sequential nature of a continuous video of a scene where the background model does not change discontinuously and can therefore be obtained by updating the background model learned from preceding images.

The above approaches mainly focus on constructing the background models from the intensity of pixels. There is also considerable research carried out to investigate new features to assist in background subtraction. In \cite{heikkila2006texture}, local binary patterns (LBP) \cite{ojala2002multiresolution} was used to formulate the background model in order to consider the local texture. In \cite{st2014flexible}, the intensity and LBP are used together to formulate the background model based on PBAS \cite{hofmann2012background}. In \cite{han2012density}, multiple features, including intensity, gradient, and Haar-like features, are combined for the foreground detection. In \cite{yang2017background}, the SILTP texture (an extension of LBP) in addition to the gray and color features is used. However, these kinds of features are often effective for a certain type of texture in a small predefined region that the current pixel is located in. When the current pixel is located in a relatively flat or poor texture region, it may fail and thus the background subtraction based on these features may not work properly as desired. 

Other than the above models, there are also methods based on support vector machines and other approaches which can be found in several survey papers \cite{piccardi2004background, radke2005image, bouwmans2008background, bouwmans2009subspace, cristani2010background, bouwmans2011recent, bouwmans2014robust, sobral2014comprehensive, bouwmans2014traditional, jeeva2015survey, xu2016background, shahbaz2017recent, bouwmans2017decomposition} as mentioned in the beginning of this Section. 

\subsection{Transform Domain Background Models}
Another class of background models are the transform domain models where the background formulation and/or foreground detection is performed in a different domain. Methods using different transformations have been proposed in the literature, including the wavelet transform, the fast Fourier transform \cite{wren2005waviz}, the discrete cosine transform \cite{porikli2005change, wang2005modeling}, the Walsh transform \cite{tezuka2008precise} and the Hadamard transform \cite{baltieri2010fast}. Here we focus on reviewing the methods using wavelet transform.

Wavelet transform has been widely used in computer vision tasks as it may capture some desired characteristics in the frequency domain. It decomposes the signal into different levels of wavelet frequency bands, which allows for multi-resolutional analysis. There have been a few methods in the literature using the wavelet transform for background subtraction. In \cite{huang2003wavelet}, a moving object segmentation method was developed in the wavelet domain, where the change detection method with different thresholds is used in four wavelet bands. The canny edge operator is applied on the results to get the edge maps of the bands. The edge result of the current frame is obtained as the union of different edge maps or the inverse transform of the different edge maps, and the foreground is detected as the video object planes based on the edges. This was further improved in \cite{huang2004double} by using a double edge detection method. In \cite{khare2015moving}, the Daubechies complex wavelet transform is used and the foreground edges are obtained using the double change detection method similarly as in \cite{huang2004double}. The final detection results are obtained by post-processing the edge map with some morphological operations. This method was further imporved in \cite{kushwaha2014complex}, where the approximate median filter is applied to detect the frame difference. In \cite{antic2009efficient}, an undecimated wavelet transform is used and the foreground is detected in each wavelet band by frame differencing. The final foreground is obtained as the union of the detection results from all the bands. This method was further improved in \cite{crnojevic2009optimal, antic2009robust} where, instead of using frame differencing for foreground detection in each wavelet band, a modified z-score test was used. The moving foreground object is detected as outliers. The above methods generally detect the foreground based on two or three consecutive frames using techniques such as frame differencing without maintaining a background model.

In \cite{gao2008new}, the orthogonal non-separable wavelet is used where the approximation coefficient is used to construct the background model in order to filter out the noise. A running average scheme is used to maintain the background model when the background has a gradual change, while when there is a sudden change, the background is replaced by the current frame. In \cite{gao2008robust, gao2009robust, gao2009traffic}, a background model based on Marr wavelet kernel was proposed and the foreground was detected in each sub-band based on binary discrete wavelet transforms. Specifically, the background model keeps a sample of intensity values for each pixel in the image and uses this sample to estimate the probability density function of the pixel intensity. The density function is estimated using the Marr wavelet kernel density estimation technique. In \cite{jalal2011robust, jalal2014framework}, the Daubechies Complex Wavelet Transform was used and the background model is formulated based on the low frequency wavelet bands. The foreground is determined by comparing the difference between the current frame and the background model. In \cite{biswas2011background}, the background model is first formulated using the low frequency wavelet bands in the discrete wavelet transform, then the inverse transform is used to obtain the background model in the image domain. The foreground is detected by comparing the difference between the background and the current frame. In \cite{hsia2014efficient}, a modified directional lifting-based 9/7 discrete wavelet transform structure was proposed in order to reduce the computation and preserve fine shape information in low resolution images. The low frequency bands are used for background modelling and the double change detection method is used to detect the foreground. In \cite{kushwaha2015framework}, the high frequency wavelet bands are used as a feature for foreground decision and the low frequency wavelet bands are used for the background modelling with an improved GMM model. The above methods generally construct the background model based on the low frequency wavelet bands to reduce the effect of the noise.

In \cite{toreyin2005moving}, a foreground detection method was proposed to process wavelet compressed video. The background model and foreground detection is implemented for each wavelet band. A running average scheme is used to formulate the background and the foreground in each band is determined by comparing the difference between the current band image and the background. Finally, the foreground of the image is detected as the union of the detected foreground in all wavelet bands. In \cite{guan2008wavelet}, the HSV components are used as features and the value component is used for foreground detection while the saturation component is used to suppress moving shadows. The foreground is detected by comparing the difference between the current frame and the background model, and the threshold used is determined based on the standard deviation of the difference image after the wavelet transform. The region connectivity with chromatic consistency is further used in the background update to overcome the aperture problem in \cite{guan2010spatio}. This method was further improved in \cite{guan2009adaptive}, where the RGB color space is used and the foreground segmentation is further developed based on the motion variation and chromatic characteristics. In \cite{guan2010motion, guan2012unsupervised}, a color ratio difference was further adopted to suppress shadow. The foreground classification is further developed in \cite{guan2014motion} by fusing mode filtering, connectivity analysis and spatial-temporal correlation. In \cite{mendizabal2011region}, a region based background model was proposed where the background for each region is modelled with GMM in the wavelet domain. The distance between the current frame and the background model is calculated as the summation of the distance in different wavelet bands. Accordingly, the background model is updated with the new distance and so is the foreground detection. The above methods generally construct the background models in all wavelet bands or construct one background model based on all the wavelet bands.

The existing wavelet transform based methods generally only take advantage of the noise resilience capability of the transform in order to construct a reliable background model or detect reliable foreground to lessen the effect of noise. Generally only one or two levels of wavelet decomposition are employed. The foreground result is usually obtained as the union of the results in different bands. Compared with these methods, our approach targets the foreground detection in camouflaged scenes where the wavelet transform is used to reveal detailed differences in different frequencies. Usually multiple levels ($6$ for example) of wavelet transform are required to show the small differences in the camouflaged scene as demonstrated in the next Section. Also we propose a fusion scheme to efficiently combine the results from different bands based on the characteristics of the wavelet transform.

\subsection{Background Models for Camouflaged Scenes}
\begin{figure}[tbp]
\begin{center}
\includegraphics[height = 0.4\hsize]{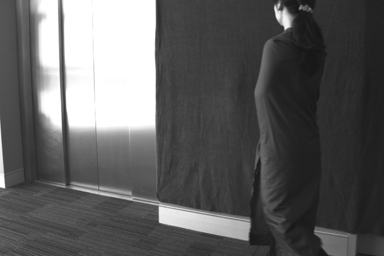}
\end{center}
\caption{Example of a camouflaged scene.}
\label{CamExample}
\end{figure}

The foreground detection methods discussed above are developed to deal with the general foreground objects which show distinct intensity or texture changes. However, there are situations where the foreground objects may share similar intensity and texture as the background, such as the camouflaged foreground, especially those with poor texture as shown in Fig. \ref{CamExample}. These cases pose great challenges to foreground detection and needless to say, methods that can properly handle such cases are highly desirable for robust foreground detection. Currently there are only a few methods developed for such cases. Here we only review the existing methods specifically designed for the camouflaged cases. Note that some methods working on regular scenes may also work in the camouflaged cases to some extent, but only the methods with a specific component designed for the camouflaged cases are described in this Subsection.

In \cite{conte2010algorithm}, a post-processing phase was used after foreground detection to recover camouflaged errors based on object detection. The paper focused on the detection of the moving people and the method is limited by the performance of person detection. Similarly in \cite{liu2012foreground}, the object information is also used and formulated over time using a Markov random field model, which assists the detection of the camouflaged part of the object in the following frames.  In \cite{zhang2016bayesian}, a camouflage model was proposed, which formulates spatial global models on the foreground and background in addition to the pixel models over time. It can also be regarded as building a model of the foreground object. All these methods require part of the object always being non-camouflaged or the object being non-camouflaged when the object first moves into the scene, in order to detect the object or formulate the global foreground model. These constraints limit their use in practical applications. By contrast, the method proposed in this paper detects the camouflaged region by highlighting the small differences using wavelet transform.

\begin{figure}[t]
\begin{center}
\begin{tabular}{c@{}c}
\includegraphics[width=0.48\hsize]{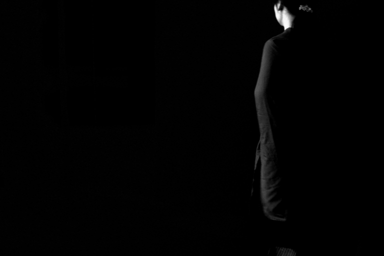} & \includegraphics[width=0.48\hsize]{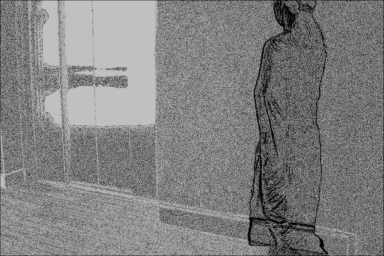}\\
(a) & (b)\\
\end{tabular}
\end{center}
\caption{Differences of (a) intensity and (b) LBP in the image domain between the current image and the background image.} 
\label{ImageDiff}
\end{figure}
\newcommand\mhwd{0.24}
\begin{figure}[tbp]
\begin{center}
\begin{tabular}{c@{}c@{}c@{}c@{}}
\includegraphics[width=\mhwd\hsize]{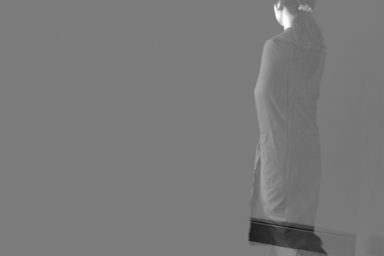} & \includegraphics[width=\mhwd\hsize]{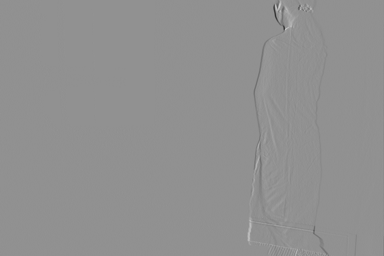}& \includegraphics[width=\mhwd\hsize]{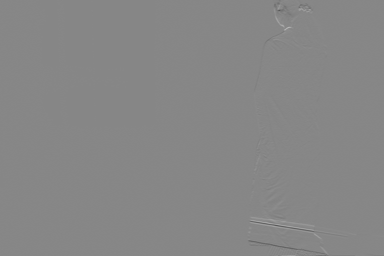}& \includegraphics[width=\mhwd\hsize]{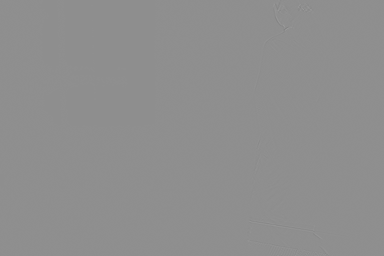}\\
\includegraphics[width=\mhwd\hsize]{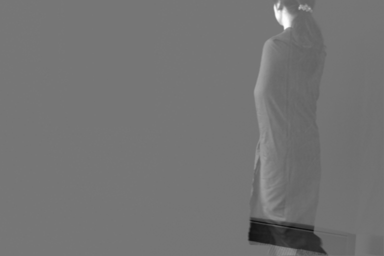} & \includegraphics[width=\mhwd\hsize]{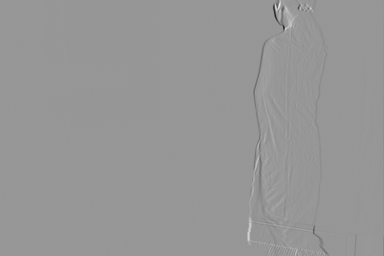}& \includegraphics[width=\mhwd\hsize]{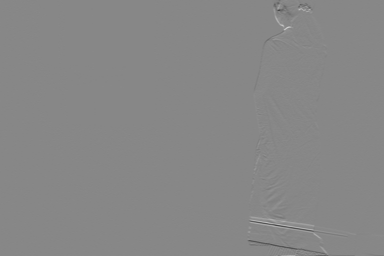}& \includegraphics[width=\mhwd\hsize]{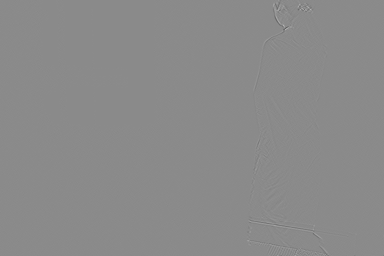}\\
\includegraphics[width=\mhwd\hsize]{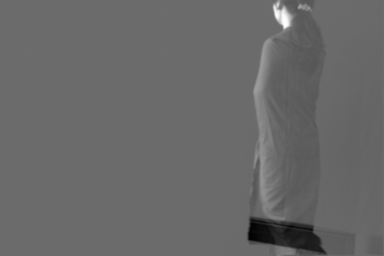} & \includegraphics[width=\mhwd\hsize]{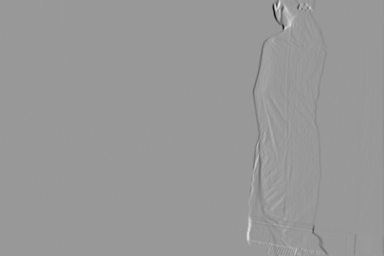}& \includegraphics[width=\mhwd\hsize]{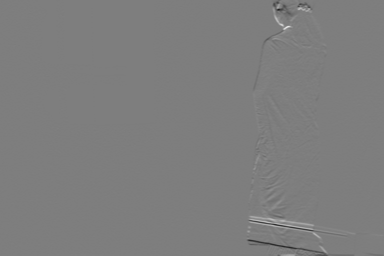}& \includegraphics[width=\mhwd\hsize]{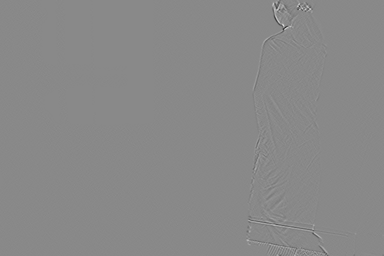}\\
\includegraphics[width=\mhwd\hsize]{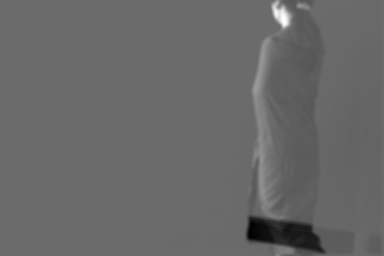} & \includegraphics[width=\mhwd\hsize]{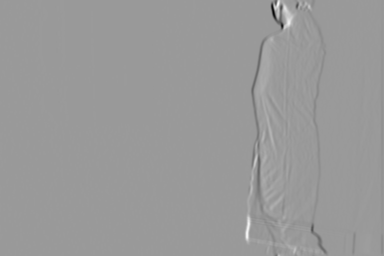}& \includegraphics[width=\mhwd\hsize]{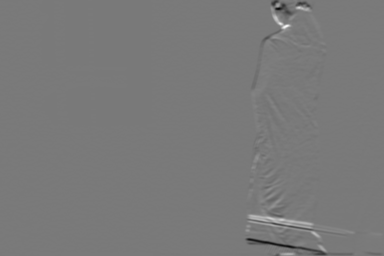}& \includegraphics[width=\mhwd\hsize]{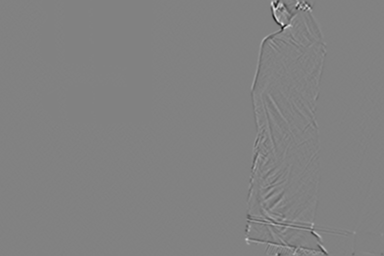}\\
\includegraphics[width=\mhwd\hsize]{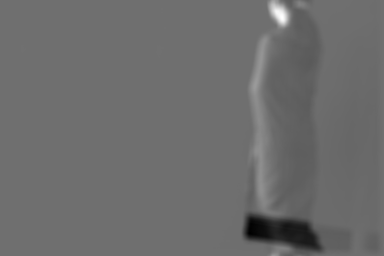} & \includegraphics[width=\mhwd\hsize]{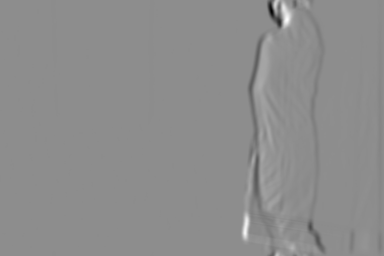}& \includegraphics[width=\mhwd\hsize]{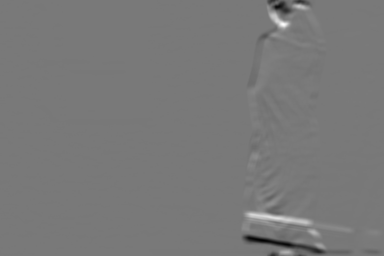}& \includegraphics[width=\mhwd\hsize]{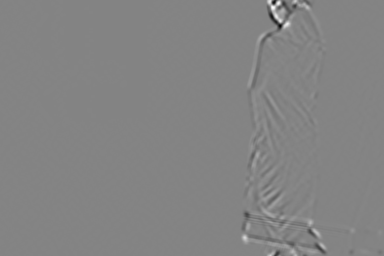}\\
\includegraphics[width=\mhwd\hsize]{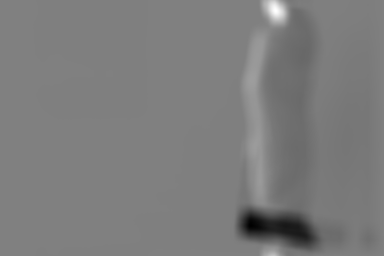} & \includegraphics[width=\mhwd\hsize]{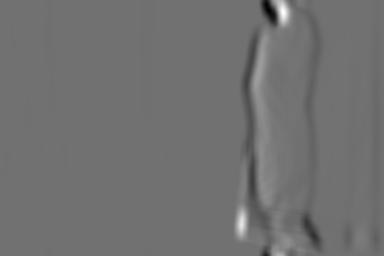}& \includegraphics[width=\mhwd\hsize]{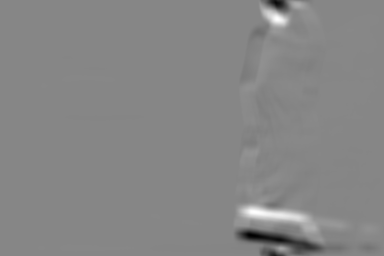}& \includegraphics[width=\mhwd\hsize]{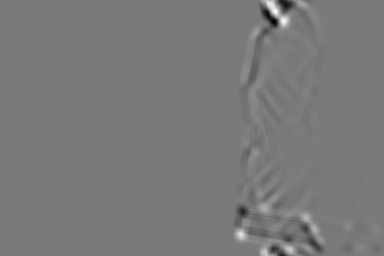}\\
\includegraphics[width=\mhwd\hsize]{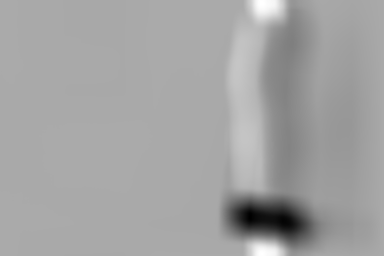} & \includegraphics[width=\mhwd\hsize]{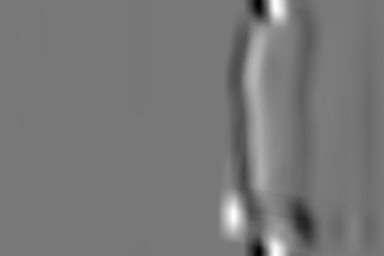}& \includegraphics[width=\mhwd\hsize]{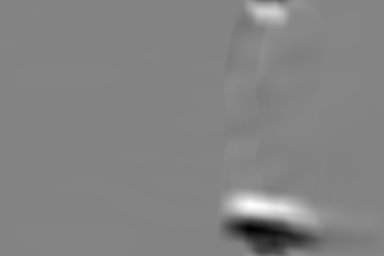}& \includegraphics[width=\mhwd\hsize]{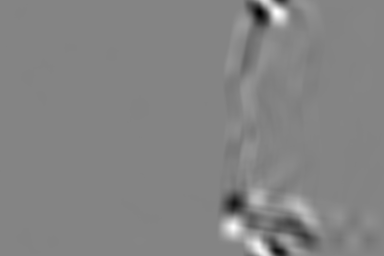}\\
LL Band & LH Band &HL Band&HH Band\\
\end{tabular}
\end{center}
\caption{Differences of wavelet coefficients between the current image and the background image in each wavelet band at different decomposition levels.} 
\label{WaveletDiff}
\end{figure}

\section{Motivation and overview}
\label{motandove}
\subsection{Motivation}
\label{sec_motivation}
Most of the background subtraction methods work on the cases where foreground object and background show distinct changes in either intensity or texture. However, as mentioned in the above section, there are camouflage cases such as Fig. \ref{CamExample} where the foreground objects share similar color to the background. Fig. \ref{ImageDiff} shows an example of differences in intensity and texture (measured by LBP) between the foreground and background (obtained by MOG2) where the values in the image are properly scaled for display. It can be seen that the differences in the visually similar regions are very small. Consequently the existing methods based on intensity or LBP features may fail in such cases.

\begin{figure}[t]
\centering
\includegraphics[width=0.9\hsize]{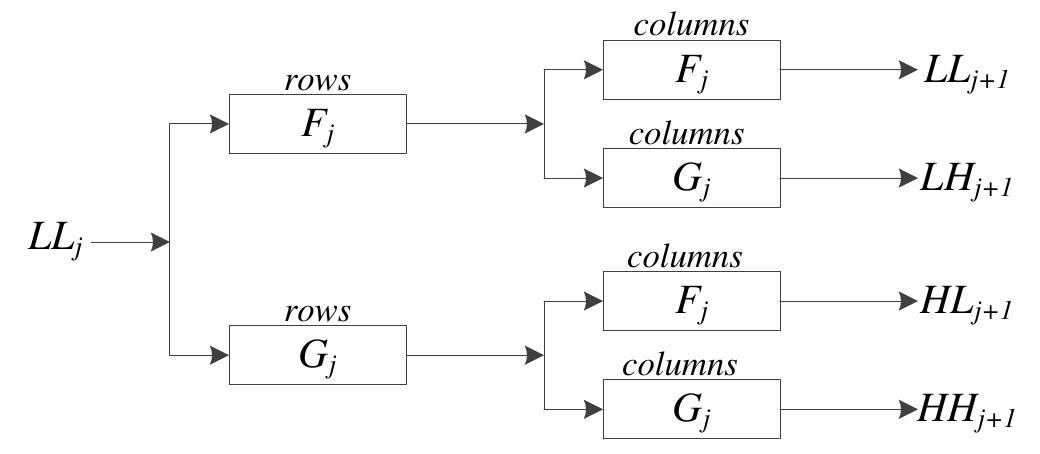}
\caption{Block diagram of the stationary wavelet transform.}
\label{fig_swt}
\end{figure}

By decomposing the signal into different bands, a wavelet transform can capture and separate different characteristics of the signal into each band. The key idea of this paper is that for different parts of a true foreground there is a detectable difference between the foreground and background in one or more bands. There are many types of wavelets including non-redundant and redundant transforms using Haar wavelet, Daubechies wavelet, etc. In this paper, we employ the stationary wavelet transform (translation-invariant) \cite{nason1995stationary, pesquet1996time} based on the Haar wavelet due to its simplicity. The stationary wavelet transform is a non-decimated redundant wavelet transform which contains the coefficients of the $\epsilon$-decimated wavelet transform for every choice of $\epsilon$. Since it is non-decimated, each wavelet band image is of the same size as the original image. Fig. \ref{fig_swt} shows the block diagram of the 2D stationary wavelet transform. $LL_j$, $LH_j$, $HL_j$ and $HH_j$ represent the low frequency approximation of the signal and the details (high frequency) of the signal in horizontal, vertical, and diagonal directions at level $j$, respectively. $F_j$ and $G_j$ represent the low-pass and high-pass filters, respectively. As shown in Fig. \ref{fig_swt}, the LL band (or the signal if it is the first decomposition level) is first filtered along rows using the low-pass and high-pass filters and then filtered again along columns. Compared with the discrete wavelet transform, there is no decimation (downsample) after the convolution with the filters in the stationary wavelet transform.

Fig. \ref{WaveletDiff} shows the difference of wavelet coefficients in each band at different levels between the current frame and background model. It can be seen that although no single band can perfectly extract the entire foreground person, different parts of the person are able to be extracted by a few and different wavelet bands. For example as shown in Fig. \ref{WaveletDiff}, the LH bands seem to strongly respond to the changes in the body of the person. Compared to the intensity and texture differences in the image domain as shown in Fig. \ref{ImageDiff}, we can see that the differences are greatly highlighted in a few wavelet bands (in relatively higher levels such as levels higher than 3), which makes the foreground detection from the visually similar background possible. Note that for cases where foreground and background are both texture-less such as rigid objects with extremely smooth surfaces, the wavelet transform may not effectively highlight the differences between them and thus they may not be detectable. For other cases where the color difference between the foreground and background is small but foreground or background are textural surfaces, such as clothes, the difference between them can become apparent or detectable in one or more wavelet bands as shown in Fig. \ref{WaveletDiff}.

\begin{figure*}[t]
\centering
\includegraphics[width=0.7\hsize]{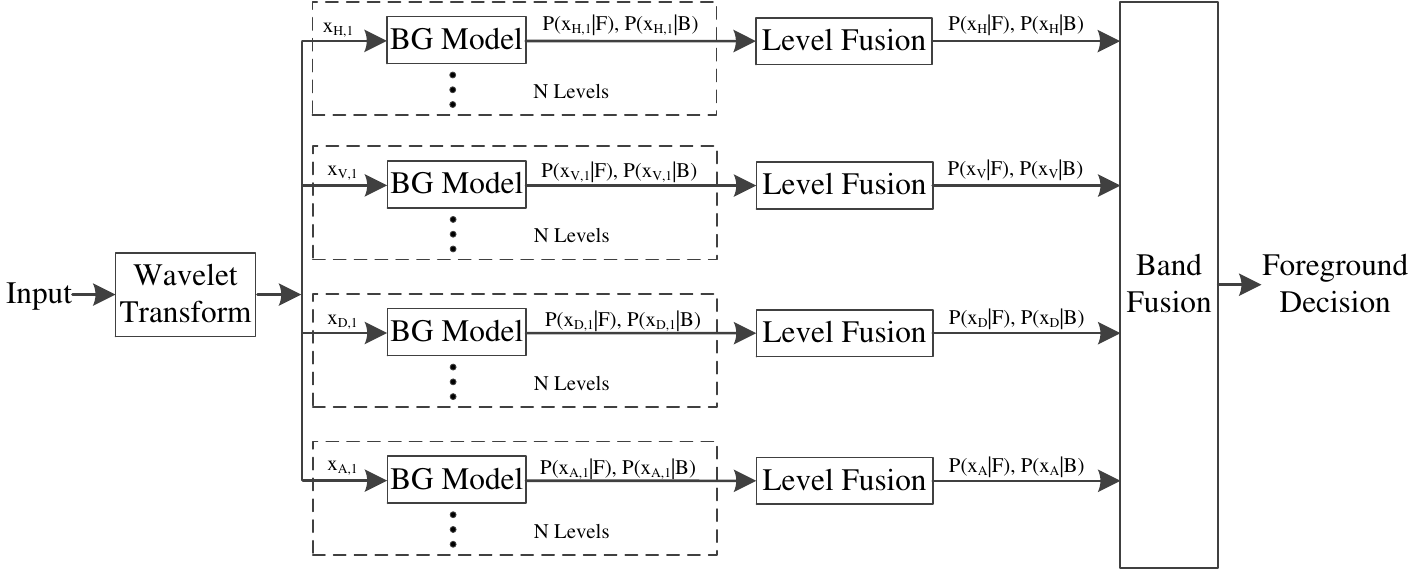}
\caption{The proposed framework for camouflaged foreground detection in the wavelet domain.}
\label{framework}
\end{figure*}

\subsection{Overview of the proposed framework}
The input image is first decomposed into different levels of wavelet bands. The foreground detection based on the wavelet bands can be considered as a decision making process from different representations. Therefore, the foreground decision can be made by
\begin{equation}
D=\left\{
\begin{aligned}
&1 \hspace{0.3cm} if ~ p(F|x_{A, N}; x_{H, 1}, ..., x_{H, N}; x_{V, 1}, ..., x_{V, N};\\ &\hspace{0.7cm} x_{D, 1}, ..., x_{D, N}) >p(B|x_{A, N}; x_{H, 1}, ..., \\& \hspace{0.7cm} x_{H, N}; x_{V, 1}, ..., x_{V, N}; x_{D, 1}, ..., x_{D, N})\\
&0 \hspace{0.3cm} else
\end{aligned}
\right.
\label{ofgdecision}
\end{equation} 
where $x_{d, l}$ represents the wavelet coefficient of the band $d$ at level $l$, while $A$, $H$, $V$ and $D$ represent the ``LL'', ``LH'', ``HL'' and ``HH'' bands, respectively. $p(F|x_{A, N}; x_{H, 1}, ..., x_{H, N}; x_{V, 1}, ..., x_{V, N}; x_{D, 1}, ..., x_{D, N})$
and $p(B|x_{A, N}; x_{H, 1}, ..., x_{H, N}; x_{V, 1}, ..., x_{V, N}; x_{D, 1}, ..., x_{D, N})$ represent the probabilities of the pixel belonging to foreground and background, respectively. $N$ is the decomposition level used in the wavelet transform. The pixel is detected as foreground ($F$) when $D=1$ and background ($B$) otherwise. 

According to the Bayesian theorem, the probabilities of the pixel belonging to foreground and background can be further represented as
\begin{small}
\begin{align}
&p(F|x_{A, N}; x_{H, 1}, ..., x_{H, N}; x_{V, 1}, ..., x_{V, N}; x_{D, 1}, ..., x_{D, N})= \nonumber\\
&\frac{p(x_{A, N}; x_{H, 1}, ..., x_{H, N}; x_{V, 1}, ..., x_{V, N}; x_{D, 1}, ..., x_{D, N}|F)P(F)}{p(x_{A, N}; x_{H, 1}, ..., x_{H, N}; x_{V, 1}, ..., x_{V, N}; x_{D, 1}, ..., x_{HH, N})} \nonumber\\[0.2cm]
&p(B|x_{A, N}; x_{H, 1}, ..., x_{H, N}; x_{V, 1}, ..., x_{V, N}; x_{D, 1}, ..., x_{D, N})= \nonumber\\
&\frac{p(x_{A, N}; x_{H, 1}, ..., x_{H, N}; x_{V, 1}, ..., x_{V, N}; x_{V, 1}, ..., x_{V, N}|B)P(B)}{p(x_{A, N}; x_{H, 1}, ..., x_{H, N}; x_{V, 1}, ..., x_{V, N}; x_{D, 1}, ..., x_{D, N})} 
\label{bayesformulation}
\end{align}
\end{small} 
Foreground can appear in the scene at any time and any location, and the foreground typically occupies less than 50\% of the image. However, for simplicity (and as commonly adopted in the literature), we assume the prior foreground and background probabilities to be the same, i.e., $p(F)=p(B)$. Therefore, with (\ref{bayesformulation}), (\ref{ofgdecision}) can be simplified as 
\begin{equation}
D=\left\{
\begin{aligned}
&1 \hspace{0.3cm} if ~ p(x_{A, N}; x_{H, 1}, ..., x_{H, N}; x_{V, 1}, ..., x_{V, N};\\  
&\hspace{1cm} x_{D, 1}, ..., x_{D, N}|F) >p(x_{A, N}; x_{H, 1}, ..., \\ & \hspace{1cm} x_{H, N}; 
x_{V, 1}, ..., x_{V, N}; x_{D, 1}, ..., x_{D, N}|B)\\
&0 \hspace{0.3cm} else
\end{aligned}
\right.
\label{fgdecision}
\end{equation}

It can be seen that in order to make the foreground decisions, the likelihood functions over all the wavelet bands need to be obtained first. 

Since the ``LL'', ``LH'', ``HL'', ``HH'' bands capture different characteristics of the signal, they can be regarded as independent from each other. Thus the following can be obtained.
\begin{align}
&p(x_{A, N}; x_{H, 1}, ..., x_{H, N}; x_{V, 1}, ..., x_{V, N}; x_{D, 1}, ..., x_{D, N}|F) \nonumber\\
& = p(x_{A, N}|F)\cdot p(x_{H, 1}, ..., x_{H, N}|F)\cdot p(x_{V, 1}, ..., x_{V, N}|F)\nonumber\\
&\hspace{0.3cm}\cdot p(x_{D, 1}, ..., x_{D, N}|F)\nonumber\\
&p(x_{A, N}; x_{H, 1}, ..., x_{H, N}; x_{V, 1}, ..., x_{V, N}; x_{D, 1}, ..., x_{D, N}|B) \nonumber\\
& = p(x_{A, N}|B)\cdot p(x_{H, 1}, ..., x_{H, N}|B)\cdot p(x_{V, 1}, ..., x_{V, N}|B)\nonumber\\
&\hspace{0.3cm}\cdot p(x_{D, 1}, ..., x_{D, N}|B)
\label{indecomp}
\end{align}
The foreground detection problem is now reduced to the estimation of the likelihood functions for each type of wavelet bands, which will be shown in the next Section. Considering that the wavelet bands are used as features instead of reconstructing the image, in this paper, all the ``LL'' wavelet bands from level 1 to level N are used.

The framework of the proposed method is shown in Fig. \ref{framework}. Wavelet transform is first applied to decompose the input image into different levels of wavelet bands, and background models are formulated for each band. Instead of making foreground decisions for each wavelet band, the likelihood functions of the coefficients being foreground and background are estimated, respectively. Considering the properties of the wavelet transform, the likelihood values calculated on each type of band are fused over different levels, shown as ``level fusion'' in Fig. \ref{framework}, and then the results of different types of bands are fused, shown as ``band fusion'' in Fig. \ref{framework}. The final foreground decision is determined based on the fused result. In this paper, we only consider foreground detection using the grayscale video and thus only one component is used. It can be easily extended to color images.

\section{The proposed method}
\label{propmethod}
\subsection{Foreground and background formulation for each wavelet band}
After video frames are decomposed into different wavelet bands, background and foreground models are updated for each wavelet band. In this paper, the widely used GMM model is used. In the conventional GMM model, each pixel in the image domain is formulated as a mixture of Gaussian distributions, while here the wavelet coefficients are modelled instead. Similarly as in \cite{zivkovic2006efficient}, the GMM model for the wavelet coefficients in each band over time can be represented by
\begin{equation}
p(x_{d, l})=\sum_{m=1}^{M}\pi_m N(x_{d, l};\mu_m,\delta_m ^2)
\end{equation}
where $d$ and $l$ represents the wavelet band type and level. $M$ is the number of Gaussian components, $\pi_m$, $\mu_m$, $\delta_m^2$ are the weight, mean and variance of the $m$-th Gaussian component for the wavelet band $d$ at level $l$ (where the subscript $d,l$ is omitted for simplicity). The samples may contain both foreground objects and background. Usually, if the foreground object presents in the scene, it will be captured by the distribution as a new Gaussian component with a small weight ($\pi_m<0.1$ for example). Since the Gaussian components are ordered by the magnitude of the weight, the background model can be approximated by the first $B$ largest Gaussian components, which is $p(x_{d, l}|B) \approx \sum_{m=1}^{B}\pi_m N(x_{d, l};\mu_m,\delta_m ^2)$.

For natural images, foreground objects are generally assumed to follow a uniform distribution \cite{zivkovic2004improved}, $p(x|F)=c_{F}$. According to the central limit theorem, the (weighted) summation of identically distributed random variables can be approximated by a Gaussian distribution. Since the wavelet coefficient in each band is obtained with a filter operation which can be regarded as a linear transform,  the wavelet coefficients of the foreground can be assumed to follow a Gaussian distribution as follows.
\begin{equation}
p(x_{d, l}|F)=N(x_{d, l};\mu _F,\delta _F ^2)
\end{equation}
where $\mu _F$ and $\delta _F ^2$ are the mean and variance of the foreground distribution for the wavelet band $d$ at level $l$ (where the subscript $d,l$ is removed for simplicity as above). For the high frequency wavelet bands including ``LH'', ``HL'' and ``HH'', the mean is zero due to the properties of the wavelet transform. In our experiments, the variances of different bands are estimated using sample video sequences and fixed through all the tests.

\subsection{Fusion of the likelihood from different wavelet bands}
One way of obtaining the likelihood functions for each type of wavelet band is using the Bayesian theorem in the same way as (\ref{indecomp}) by assuming they are independent from each other. Denote $p(x_{d, 1}, ..., x_{d, N}|F)$ and $p(x_{d, 1}, ..., x_{d, N}|B)$ by $p(x_{d}|F)$ and $p(x_{d}|B)$, respectively. The likelihood functions for one type of wavelet bands can be obtained as follows.
\begin{align}
p(x_{d}|F)=p(x_{d, 1}|F) p(x_{d, 2}|F) \ldots p(x_{d, N}|F) \nonumber\\
p(x_{d}|B)=p(x_{d, 1}|B) p(x_{d, 2}|B) \ldots p(x_{d, N}|B)
\end{align}
However, one type of wavelet bands over different levels all capture details in the same direction and thus are related to each other. Therefore, the above decomposition may not be able to well represent the likelihood functions. In the following, we propose a weighted fusion way to obtain the likelihood functions for each type of wavelet bands.

It is known that each wavelet coefficient is obtained based on a number of pixels in a block and the block size increases as the level increases. Thus the result obtained for the wavelet coefficient from each band is a noisy result corresponding to a block.  Therefore, from the perspective of noise reduction, the likelihood functions from one type of wavelet bands can be obtained by 
\begin{align}
p(x_{d}|F)=\frac{1}{N}\sum_{l=1}^{N}f(p(x_{d,l}|F)) \nonumber\\
p(x_{d}|B)=\frac{1}{N}\sum_{l=1}^{N}f(p(x_{d,l}|B))
\label{sumfusion}
\end{align}
where $f(\cdot)$ maps the result from the wavelet domain to the image domain. In this paper, a linear mapping process is used, which is shown in the following.

It is clear that if all the pixels related to a coefficient belong to the same object, the result obtained based on this coefficient can well represent the result on these pixels. On the contrary, if pixels related to a coefficient belong to different objects, the result obtained based on this coefficient may not be correct for all these pixels. In order to characterize this relationship, the correlation among pixels needs to be modelled first. It is known that the image signal can be modelled as a first-order autoregressive process $x(i)=c + \alpha \cdot x(i-1) + \epsilon$, where $\alpha$ ($0<\alpha<1$) is the autoregressive coefficient, $c$ is a constant, and $\epsilon$ is a white noise process with zero mean. Based on this model, in this paper, the correlation among pixels is simply represented as
\begin{equation}
\rho (\delta) = \alpha ^{\delta}
\end{equation}
where $\delta$ represents the distance between two pixels. When the distance between two pixels increases, their correlation ($\rho$) reduces. Therefore, as the wavelet decomposition level increases and accordingly the number of pixels related to one coefficient increases, more pixels are becoming far away from the central pixel. Consequently, the correlation between these pixels and the central pixel gets smaller, and the result obtained based on the coefficient is becoming less representative for all its related pixels.

To account for this difference among different levels, a translation weight $\omega _{t,l}$ is introduced. It is determined as the average correlation of all the pixels related to the coefficient.  

\begin{equation}
\omega _{t,l}=\frac{1}{N_p}\sum_{i=1}^{N_p} {\rho(\delta_i)}
\label{translationweight}
\end{equation}
where $N_p$ is the total number of the pixels related to one coefficient in the wavelet bands at level $l$ and $\delta_i$ represents the distance between the pixel and the center of the block.

On the other hand, noise exists in images and thus in the wavelet bands. The results obtained at each level are all affected by the noise to some extent. Assume it is white Gaussian noise, which has zero mean and the same energy at different frequencies. Since the energies of the signal at different wavelet bands of different levels can be quite different, the effects of the noise on the results of different bands are different. To take this into account, a noise-induced weight factor is introduced as follows.
\begin{small}
\begin{equation}
\omega _{n,d,l} =\frac{\sigma _{d,l} - \sigma _n}{\sigma _{d,l}}
\label{noise_weight}
\end{equation}
\end{small}
where $\sigma _{d,l}$ and $\sigma _n$ are the standard deviation of the wavelet band of type $d$ at level $l$ and noise, respectively.

The mapping function $f(\cdot)$ in (\ref{sumfusion}) is defined by combining the translation weight in (\ref{translationweight}) and the noise induced weight in (\ref{noise_weight}) as $f(p(x_{d,l}|\cdot))=\omega _{d,l}p(x_{d,l}|\cdot)$, where $\omega _{d,l}=\omega _{t,l} \cdot \omega _{n,d,l}$. Accordingly the likelihood functions for each wavelet band in (\ref{sumfusion}) can be obtained as
\begin{align}
p(x_{d}|F)=\frac{1}{N}\sum_{l=1}^{N}\omega _{d,l}p(x_{d,l}|F) \nonumber\\
p(x_{d}|B)=\frac{1}{N}\sum_{l=1}^{N}\omega _{d,l}p(x_{d,l}|B)
\end{align}

The above process fuses the results from one type of wavelet bands among different levels. Then according to (\ref{indecomp}), the result from different types of wavelet bands can be further fused. Accordingly, the foreground decisions can be made via (\ref{fgdecision}).

In the implementation, since the ``LL'' wavelet bands focus on the intensity while the other bands focus on the texture, the ``LL'' wavelet bands are processed separately and the final foreground detection result is obtained by combining the fused result from the high frequency wavelet bands and the result from the low frequency bands. Also for the ``LL'' wavelet bands, the foreground distribution is regarded as a uniform distribution in the same way as in the conventional GMM model for simplicity.

\section{Experiments}
\label{experimentsec}
\subsection{Experimental setting}
Since we focus on the camouflaged foreground detection problems, camouflaged videos are used for evaluation. The camouflaged videos in \cite{brutzer2011evaluation} and SBM-RGBD dataset \cite{Camplani2017RGBD} are both used in our experiments. The video in \cite{brutzer2011evaluation} was artificially generated by computers and the 5 videos in \cite{Camplani2017RGBD} are recorded as RGB-D dataset using the Microsoft Kinect. In the experiments, the depth channel is not considered. Moreover, we further recorded 10 videos captured in real scenes including both in-house and out-of-house cases. The dataset is named as ``CAMO\_UOW'' and the details of the dataset including the resolution, number of frames and the format (grayscale or RGB) are shown in Table \ref{detail_CAMO_UOW}. In our recorded sequences, the foreground person wears clothes in a similar color as that of the background. Groundtruth are manually labelled for all frames in the collected ``CAMO\_UOW'' \footnote{CAMD\_UOW is available at \url{https://www.uow.edu.au/~wanqing/#Datasets}.}. Example frames of the sequences in ``CAMO\_UOW'' are shown in Fig. \ref{datasetexample} along with the groundtruth foreground masks.

\begin{table}[tbp]
\centering
\caption{Details on the collected camouflaged video dataset (CAMO\_UOW). The groundtruth is available for all frames.}
\label{detail_CAMO_UOW}
\begin{tabular}{c c c c}
\hline
Video number & Resolution & Frames & Format \\
\hline
Video 1 & $1600 \times 1200$ & $371$ & Grayscale \\
\hline
Video 2 & $1600 \times 1200$ & $176$ & Grayscale \\
\hline
Video 3 & $1600 \times 1200$ & $371$ & Grayscale \\
\hline
Video 4 & $1600 \times 1200$ & $371$ & Grayscale \\
\hline
Video 5 & $1600 \times 1200$ & $371$ & Grayscale \\
\hline
Video 6 & $1600 \times 1200$ & $373$ & Grayscale \\
\hline
Video 7 & $1920 \times 1080$ & $272$ & RGB \\
\hline
Video 8 & $1920 \times 1080$ & $466$ & RGB \\
\hline
Video 9 & $1920 \times 1080$ & $288$ & RGB \\
\hline
Video 10 & $1920 \times 1080$ & $458$ & RGB \\
\hline
\end{tabular}
\end{table}

In our implementation, the images are decomposed into six levels of wavelet bands. Accordingly, the sequences are cropped into the nearest size that can be divided by $2^6$ due to the requirement of wavelet transform. Moreover, all the sequences are converted into the grayscale format which makes the foreground detection problem more difficult. Since the wavelet coefficients in each band are modelled using the GMM model, the MOG2 implementation in openCV to be specific, several parameters are involved, including the initial variance, minimum and maximum variance of the Gaussian distributions. In our implementation, the parameters are set based on those used in the MOG2 implementation, in a proportional way according to their variance. That is to say, $\sigma^2_{ini, prop}=\sigma^2_{ini, MOG2}*\sigma^2_{w}/\sigma^2_{o}$, where $\sigma^2_{ini, prop}$, $\sigma^2_{ini, MOG2}$, $\sigma^2_{w}$ and $\sigma^2_{o}$ represent the initial variance used for the proposed method of a wavelet band, the initial variance used in MOG2 for the natural image, the variance of the wavelet band and the variance of the natural image, respectively. The rest of the parameters are set in the same way. For the high frequency wavelet bands including ``LH'', ``HL'', ``HH'', when the magnitude of the coefficient deviates from zero, it usually represents edges and the Gaussian distribution used to formulate such coefficients may express a large change over time. Therefore, the initial variance, minimum and maximum variance of the Gaussian are set to be larger when the mean deviates from 0. Taking the initial variance for example, it is set as $\sigma_{ini, prop}^2+\alpha \cdot m^2$, where $m$ represents the mean of the Gaussian distribution and $\alpha$ is a small weight. In the end, the detection results are processed with some morphological operations to remove the isolated pixels. To be specific, to remove the small holes inside a foreground object, the result is first processed with a closing operation and an eroding operation. Then combining with the original result, it is further processed with a closing and opening operation to remove noise. The final result is further eroded considering that the edges of the foreground objects may get dilated due to wavelet transform.

\newcommand\mhd{0.166}
\newcommand\datasethsep{-0.2cm}
\begin{figure}[t]
 \captionsetup[subfigure]{aboveskip=-0.01pt}
\centering
\subcaptionbox{Video 1}{
 \includegraphics[width = \mhd\hsize]{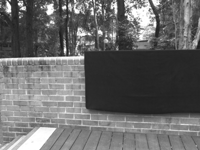}
 \hspace{\datasethsep}
 \includegraphics[width = \mhd\hsize]{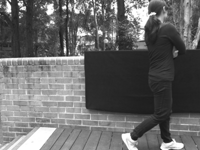}
 \hspace{\datasethsep}
 \includegraphics[width = \mhd\hsize]{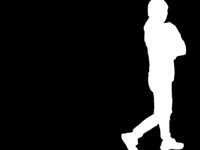}}%
\subcaptionbox{Video 2}{
 \includegraphics[width = \mhd\hsize]{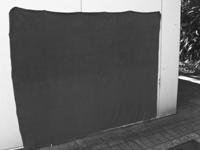}
 \hspace{\datasethsep}
 \includegraphics[width = \mhd\hsize]{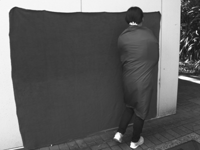}
 \hspace{\datasethsep}
 \includegraphics[width = \mhd\hsize]{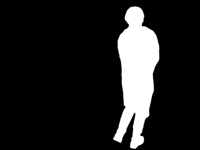}}%
 
\subcaptionbox{Video 3}{
 \includegraphics[width = \mhd\hsize]{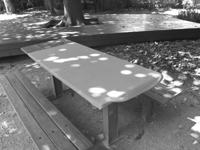}
 \hspace{\datasethsep}
 \includegraphics[width = \mhd\hsize]{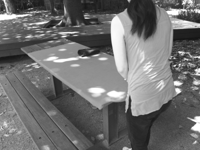}
 \hspace{\datasethsep}
 \includegraphics[width = \mhd\hsize]{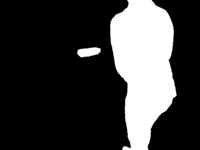}}%
\subcaptionbox{Video 4}{
 \includegraphics[width = \mhd\hsize]{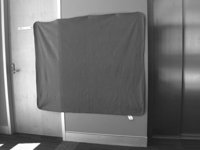}
 \hspace{\datasethsep}
 \includegraphics[width = \mhd\hsize]{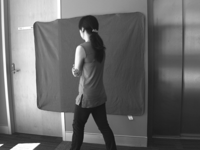}
 \hspace{\datasethsep}
 \includegraphics[width = \mhd\hsize]{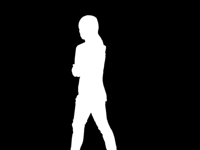}}%
 
\subcaptionbox{Video 5}{
 \includegraphics[width = \mhd\hsize]{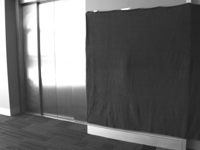}
 \hspace{\datasethsep}
 \includegraphics[width = \mhd\hsize]{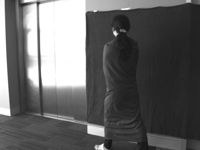}
 \hspace{\datasethsep}
 \includegraphics[width = \mhd\hsize]{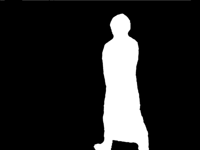}}%
\subcaptionbox{Video 6}{
 \includegraphics[width = \mhd\hsize]{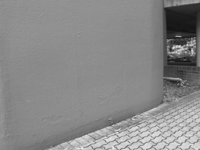}
 \hspace{\datasethsep}
 \includegraphics[width = \mhd\hsize]{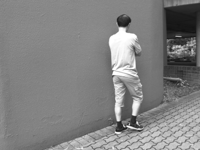}
 \hspace{\datasethsep}
 \includegraphics[width = \mhd\hsize]{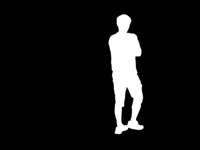}}%
 
\subcaptionbox{Video 7}{
 \includegraphics[width = \mhd\hsize]{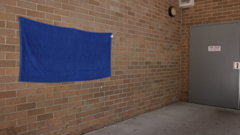}
 \hspace{\datasethsep}
 \includegraphics[width = \mhd\hsize]{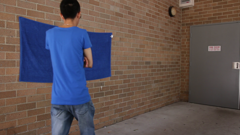}
 \hspace{\datasethsep}
 \includegraphics[width = \mhd\hsize]{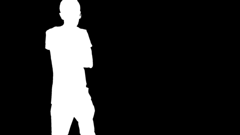}}%
\subcaptionbox{Video 8}{
 \includegraphics[width = \mhd\hsize]{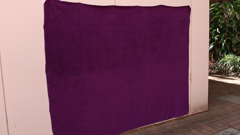}
 \hspace{\datasethsep}
 \includegraphics[width = \mhd\hsize]{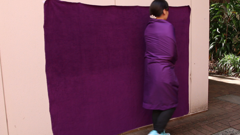}
 \hspace{\datasethsep}
 \includegraphics[width = \mhd\hsize]{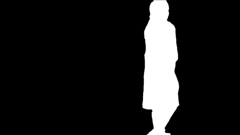}}%
 
\subcaptionbox{Video 9}{
 \includegraphics[width = \mhd\hsize]{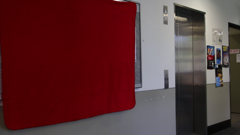}
 \hspace{\datasethsep}
 \includegraphics[width = \mhd\hsize]{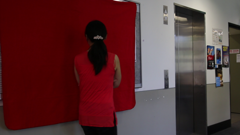}
 \hspace{\datasethsep}
 \includegraphics[width = \mhd\hsize]{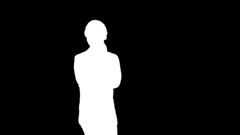}}%
\subcaptionbox{Video 10}{
 \includegraphics[width = \mhd\hsize]{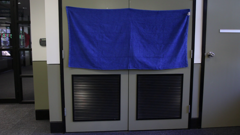}
 \hspace{\datasethsep}
 \includegraphics[width = \mhd\hsize]{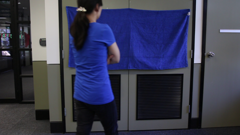}
 \hspace{\datasethsep}
 \includegraphics[width = \mhd\hsize]{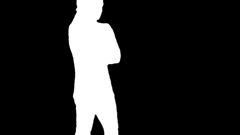}}%
\caption{Example frames and groundtruth in our collected video dataset ``CAMO\_UOW''. }
\label{datasetexample}
\end{figure}

\begin{table*}[tbp]
\centering
\setlength{\tabcolsep}{3pt}
\centering
\caption{Performance comparison on different tested videos in terms of \textit{F-measure}.}\label{aveResult}
\begin{tabular}{lccccccccc}
\hline
video & MOG2\cite{zivkovic2006efficient} & FCI\cite{el2008fuzzy} & LBA-SOM\cite{maddalena2008self} & PBAS\cite{hofmann2012background} & SuBSENSE\cite{st2015subsense} & ML-BGS\cite{yao2007multi} & DECOLOR\cite{zhou2013moving} & COROLA\cite{shakeri2016corola} & FWFC \\
\hline
Video 1   & 0.79	 & 0.88	 & 0.8	 & 0.9	 & 0.89	 & 0.89	 & 0.92	 & 0.8	 & \textbf{0.94} \\ 
\hline
Video 2   & 0.82	 & 0.79	 & 0.8	 & 0.82	 & 0.88	 & 0.8	 & 0.83	 & 0.58	 & \textbf{0.96} \\ 
\hline
Video 3   & 0.88	 & 0.86	 & 0.85	 & 0.91	 & 0.9	 & 0.8	 & 0.9	 & 0.82	 & \textbf{0.94} \\ 
\hline
Video 4   & 0.89	 & 0.9	 & 0.76	 & 0.93	 & 0.78	 & 0.88	 & \textbf{0.95}	 & 0.87	 & 0.94 \\ 
\hline
Video 5   & 0.84	 & 0.86	 & 0.82	 & 0.83	 & 0.82	 & 0.8	 & 0.82	 & 0.75	 & \textbf{0.91} \\ 
\hline
Video 6  & 0.93	 & 0.87	 & 0.77	 & 0.95	 & 0.92	 & 0.95	 & \textbf{0.97}	 & 0.72	 & 0.94 \\ 
\hline
Video 7   & 0.76	 & 0.83	 & 0.88	 & 0.91	 & 0.87	 & 0.79	 & 0.91	 & 0.83	 & \textbf{0.96} \\ 
\hline
Video 8   & 0.83	 & 0.87	 & 0.85	 & 0.87	 & 0.93	 & 0.86	 & 0.86	 & 0.68	 & \textbf{0.96} \\ 
\hline
Video 9   & 0.89	 & 0.9	 & 0.87	 & 0.84	 & \textbf{0.92}	 & 0.87	 & 0.86	 & 0.78	 & 0.88 \\ 
\hline
Video 10   & 0.89 & 0.86	 & 0.89	 & 0.91	 & 0.92	 & 0.90	 & 0.94	 & 0.85	 & \textbf{0.96} \\ 
\hline
Camouflage\cite{brutzer2011evaluation}   & 0.72	 & 0.72	 & 0.71	 & 0.82	 & 0.82	 & 0.78	 & \textbf{0.84}	 & 0.76	 & 0.75 \\ 
\hline
colorCam1\cite{Camplani2017RGBD}    & 0.15	 & 0.1	 & \textbf{0.82}	 & 0.42	 & 0.08	 & 0.09	 & 0.26	 & 0.59	 & 0.61 \\ 
\hline
colorCam2\cite{Camplani2017RGBD}    & 0.76	 & 0.22	 & 0.6	 & 0.52	 & 0.74	 & 0.28	 & 0.48	 & \textbf{0.94}	 & 0.63 \\ 
\hline
Cespatx\_ds\cite{Camplani2017RGBD}    & 0.71	 & 0.69	 & 0.86	 & 0.81	 & 0.9	 & 0.6	 & 0.74	 & 0.86	 & \textbf{0.92} \\ 
\hline
Hallway\cite{Camplani2017RGBD}    & 0.59	 & 0.62	 & 0.43	 & 0.62	 & \textbf{0.69}	 & 0.58	 & 0.67	 & 0.63	 & 0.67 \\ 
\hline
Office2\cite{Camplani2017RGBD}    & 0.33	 & 0.36	 & 0.79	 & 0.58	 & 0.48	 & 0.47	 & 0.89	 & 0.76	 & \textbf{0.9} \\ 
\hline
\textbf{Average}   & 0.74	 & 0.71	 & 0.78	 & 0.79	 & 0.78	 & 0.71	 & 0.8	 & 0.76	 & \textbf{0.87} \\ 
\hline
\end{tabular}
\end{table*}

\begin{figure*}[t]
\centering
 \includegraphics[width = 0.8\hsize]{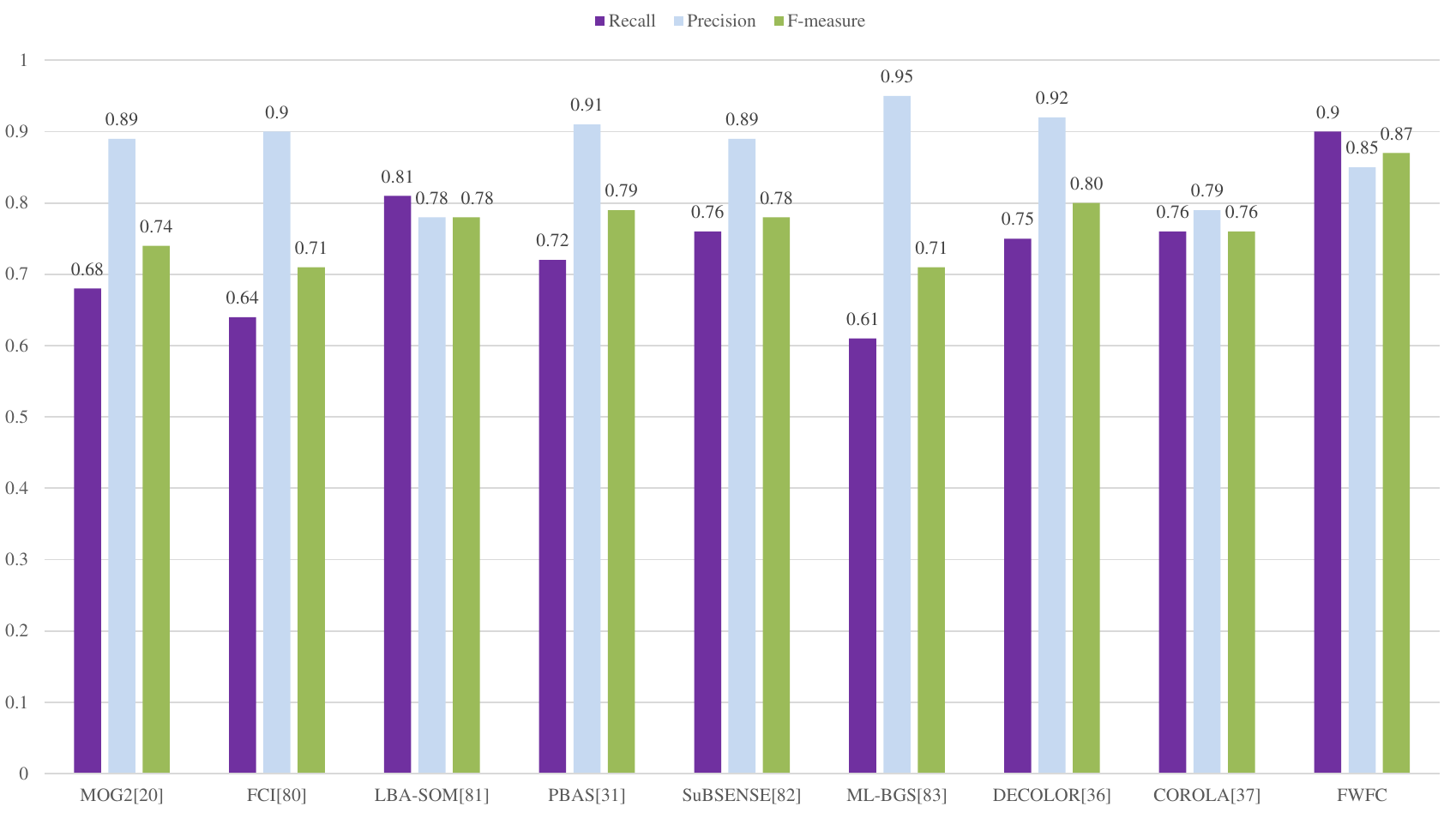}
\caption{The average performance of all the test videos in terms of \textit{Recall}, \textit{Precision} and \textit{F-measure} for different methods. }
\label{barresult}
\end{figure*}

\begin{figure*}[tbp]
\centering
 \includegraphics[width = 1\hsize]{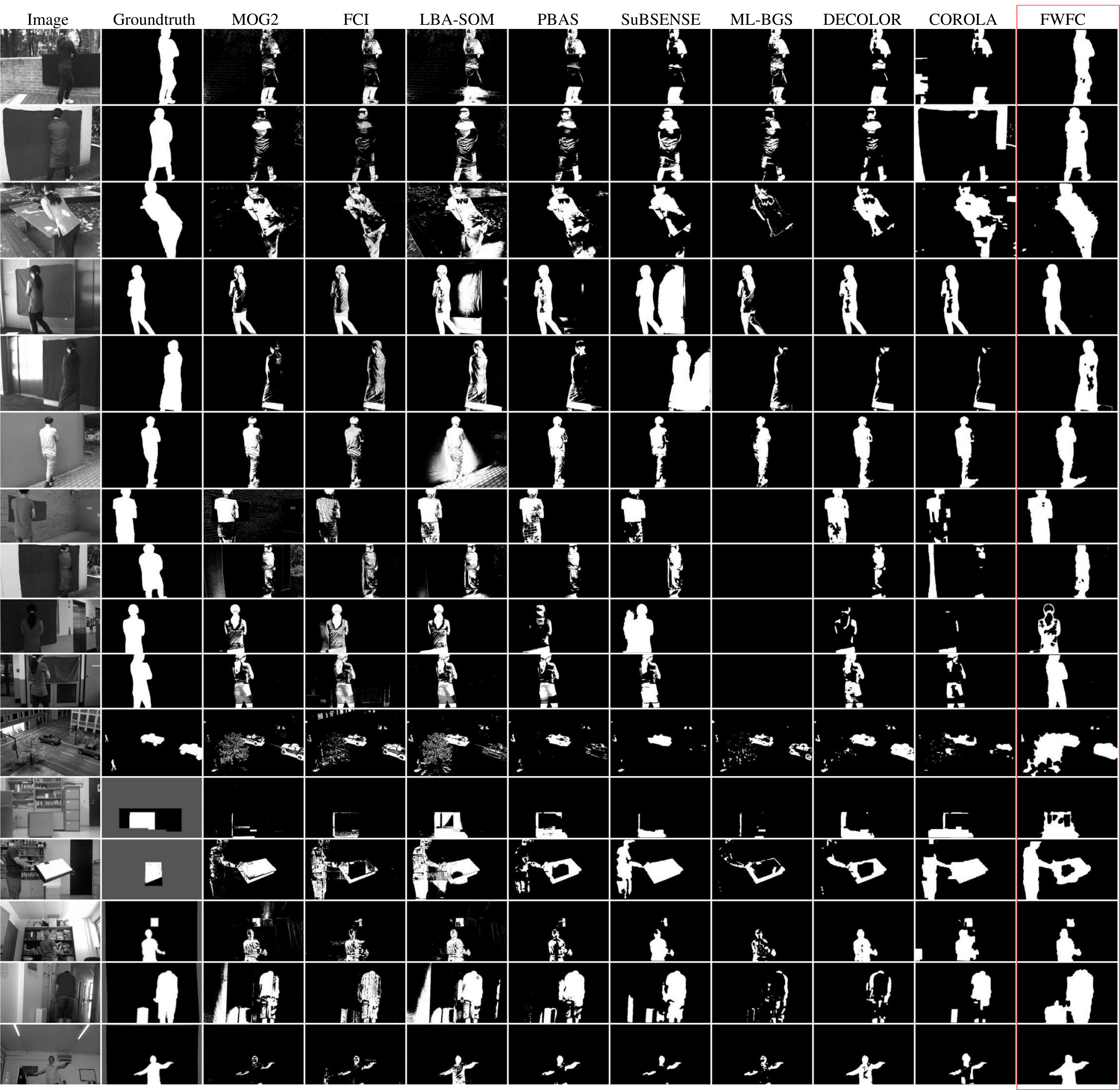}
 
\caption{Example of foreground detection results obtained using different methods on all the videos. From left to right: the original frame, the groundtruth foreground mask, and the results obtained by different methods including ``MOG2'' \cite{zivkovic2006efficient}, ``FuzzyChoquetIntegral'' \cite{el2008fuzzy}, ``LBAdaptiveSOM'' \cite{maddalena2008self}, ``PBAS'' \cite{hofmann2012background}, ``SuBSENSE'' \cite{st2015subsense}, ``MultiLayerBGS'' \cite{yao2007multi}, ``DECOLOR'' \cite{zhou2013moving}, ``COROLA'' \cite{shakeri2016corola}, the proposed FWFC. From top to bottom: Video 1-10 in our collected ``CAMO\_UOW'' dataset, the camouflaged video in \cite{brutzer2011evaluation}, colorCam1\cite{Camplani2017RGBD}, colorCam2\cite{Camplani2017RGBD}, Cespatx\_ds\cite{Camplani2017RGBD}, Hallway\cite{Camplani2017RGBD}, Office2\cite{Camplani2017RGBD}.}
\label{result_figures}
\end{figure*}

\subsection{Performance evaluation}
The performance of the proposed ``FWFC'' is shown both qualitatively and quantitatively. Several popular and representative methods are compared, including ``MOG2'' \cite{zivkovic2006efficient}, ``FuzzyChoquetIntegral'' \cite{el2008fuzzy}, ``LBAdaptiveSOM'' \cite{maddalena2008self}, ``MultiLayerBGS'' \cite{yao2007multi}, ``SuBSENSE'' \cite{st2015subsense}, ``PBAS'' \cite{hofmann2012background}, ``DECOLOR'' \cite{zhou2013moving}, ``COROLA'' \cite{shakeri2016corola}. The ``LSD'' method \cite{liu2015background} is similar to ``DECOLOR'' \cite{zhou2013moving} and ``COROLA'' \cite{shakeri2016corola} methods in terms of low-rank representation and performance on the camouflaged videos and so is not further included in the comparison. The results of these methods are obtained based on the bgslibrary \cite{bgslibrary, bgslibrarychapter} and the LRSLibrary \cite{lrslibrary2015, bouwmans2015}. The default settings in these libraries are used which is explained in the survey paper \cite{sobral2014comprehensive} and \cite{bouwmans2017decomposition}, respectively. No post-processing such as morphological operations is employed for the comparison methods since the result in our experiments shows no noticeable improvement. Note that ``DECOLOR'' \cite{zhou2013moving} cannot directly work on the sequences as the memory required is larger than 32GB and takes a very long time. Therefore, its results are first obtained on downsampled videos (the largest dimension being $320$ with the same aspect ratio) and then upsampled to its original size. There is a block-by-block implementation available in the LRSLibrary enabling the direct processing of sequences with large resolutions. However, its performance is worse than processing with downsampling and upsampling in our experiments. 

The results obtained with different methods for some example frames are shown in Fig. \ref{result_figures}. The images in each row from left to right represent the original frame, the groundtruth foreground mask, and the results obtained by different methods including ``MOG2'' \cite{zivkovic2004improved}, ``FuzzyChoquetIntegral'' \cite{el2008fuzzy}, ``LBAdaptiveSOM'' \cite{maddalena2008self}, ``PBAS'' \cite{hofmann2012background}, ``SuBSENSE'' \cite{st2015subsense}, ``MultiLayerBGS'' \cite{yao2007multi}, ``DECOLOR'' \cite{zhou2013moving}, ``COROLA'' \cite{shakeri2016corola}, the proposed FWFC. Each row of images from top to bottom represent Video 1-10 in our collected ``CAMO\_UOW'' dataset, the camouflaged video in \cite{brutzer2011evaluation}, colorCam1\cite{Camplani2017RGBD}, colorCam2\cite{Camplani2017RGBD}, Cespatx\_ds\cite{Camplani2017RGBD}, Hallway\cite{Camplani2017RGBD}, Office2\cite{Camplani2017RGBD}. From the results, it can be clearly seen that FWFC achieves the best result around the camouflaged foreground area for most of the sequences. Taking the ``Video 1'' shown in the first row of Fig. \ref{result_figures} for example, the existing methods cannot detect the body of the foreground person who wears clothes in the similar color as the background wall. On the contrary, FWFC works very well as most of the body has been detected. However, for colorCam1\cite{Camplani2017RGBD} and colorCam2\cite{Camplani2017RGBD}, part of the camouflaged foreground objects are not able to be detected. This is mainly because the foreground object and background in these two videos are both texture-less objects.

The performance is also quantitatively measured using \textit{Recall}, \textit{Precision} and \textit{F-measure}. For ``Video 1'' - ``Video 10'' in our collected dataset ``CAMO\_UOW'' and the camouflaged video in \cite{brutzer2011evaluation}, groundtruth is available for all frames and thus the \textit{Recall}, \textit{Precision} and \textit{F-measure} of the static quality metrics in \cite{vacavant2012benchmark} is used, and computed by the BMC Wizard software. For the sequences in \cite{Camplani2017RGBD}, groundtruth is only available for a few frames and the quality metrics are computed with the scripts provided by \cite{Camplani2017RGBD}. Table \ref{aveResult} shows the result of the proposed method in terms of \textit{F-measure} compared with the existing methods. ``Average'' represents the average result of all the videos. The notations of the existing methods are simplified due to the limited space, and their full notations can be found in the beginning of this Subsection. It can be seen that the performance of FWFC is much better than the existing methods, with an average \textit{F-measure} of 0.87, compared to values between 0.71 and 0.8 for the existing methods. It is worth noting that the performance on the camouflaged video in \cite{brutzer2011evaluation} is slightly worse than some of the existing methods. This is mainly because FWFC is implemented based on MOG2 \cite{zivkovic2006efficient}
which performs slightly worse in the case of swaying background such as the leaves of the tree. As shown in Fig. \ref{result_figures}, part of the trees are still detected as the foreground which is the same as the MOG2 result. 
However, it can be seen that FWFC works very well in capturing the camouflaged foreground such as the car. Also, FWFC still achieves better performance than MOG2, which FWFC is developed upon.

The average result of the tested videos in terms of \textit{Recall}, \textit{Precision} and \textit{F-measure} is shown in Fig. \ref{barresult}. It can be clearly seen that the \textit{Recall} of FWFC is much higher than others, which indicates that FWFC can detect most of the foreground objects. In terms of the \textit{Precision}, the performance of FWFC is comparable to other methods. Therefore, combined with a higher \textit{Recall}, FWFC achieves the best result. Experiments on regular (not camouflaged) videos were also conducted using the sequences from the CDW-2014 (2014 IEEE Change Detection Workshop) dataset \cite{goyette2012changedetection, changedetectionnet}. The results obtained by the proposed method are better than or comparable to those obtained by the conventional MOG2.

The computational complexity (in terms of CPU time) of FWFC in comparison with the existing methods are shown in Table \ref{result_time}. The time is for processing a frame of resolution $640*384$ and is obtained on a personal computer (i7-4790 CPU with 32GB memory). Note that ``DECOLOR'' \cite{zhou2013moving} and ``COROLA'' \cite{shakeri2016corola} are implemented in Matlab while others are implemented in C++. FWFC takes more time than MOG2 \cite{zivkovic2006efficient} but less time than most of the existing methods as listed in Table \ref{result_time}. Currently, FWFC is programmed in a sequential manner without optimization. Among the time (0.275 seconds) used to process a frame, the wavelet decomposition takes 0.12 seconds, the background modelling (MOG2) on the wavelet bands takes 0.04 seconds and the fusion from the results obtained from the wavelet bands takes around 0.11 seconds. Note that FWFC is highly parallelable considering the result from each wavelet band can be obtained in parallel, which can further reduce the time.

\begin{table}[tbp]
\centering
\caption{Complexity comparison in terms of running time (seconds) for one frame of resolution 640*384.}\label{result_time}
\begin{tabular}{l @{\hskip 4cm} c}
\hline
Methods & Time \\
\hline
MOG2\cite{zivkovic2006efficient}  & 0.007\\
FCI\cite{el2008fuzzy}  & 0.251\\
LBA-SOM\cite{maddalena2008self}  & 0.076\\
PBAS\cite{hofmann2012background}  & 0.449\\
SuBSENSE\cite{st2015subsense}  & 1.176\\
ML-BGS\cite{yao2007multi}  & 0.293\\
DECOLOR\cite{zhou2013moving}  & 4.999\\
COROLA\cite{shakeri2016corola}  & 0.619\\
\textbf{FWFC} & 0.275\\
\hline
\end{tabular}
\end{table}

\section{Conclusion}
\label{concsec}
This paper presents a fusion framework to address the camouflaged moving foreground detection problem. It first transfers the foreground detection problem into the wavelet domain and shows that the small differences in the image domain can be detected in certain wavelet bands. Then foreground and background models are formulated for all the wavelet bands. The results from different bands are fused by considering the properties of different wavelet bands. Experimental results have shown that the proposed method performs significantly better than the existing methods in terms of the camouflaged foreground detection.

\bibliographystyle{IEEEtran}
\bibliography{Reference}

\begin{IEEEbiography}[{\includegraphics[width=1in,height=1.25in,clip,keepaspectratio]{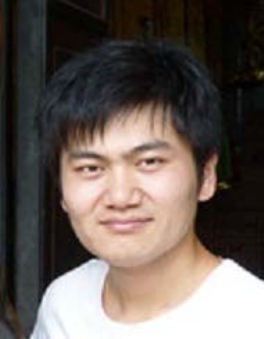}}]{Shuai Li}
is pursuing the Ph.D. degree in computer science with the University of Wollongong, Australia. He received the B.Eng. degree from Yantai University, China, in 2011, and the M.Eng. degree from Tianjin University, China, in 2014. He was with the University of Electronic Science and Technology of China as a Research Assistant from 2014 to 2015. His research interests include image/video coding, 3D video processing, and computer vision. He was a co-recipient of two best paper awards at the IEEE BMSB 2014 and IIH-MSP 2013, respectively.
\end{IEEEbiography}
\vspace{-9 mm}

\begin{IEEEbiography}[{\includegraphics[width=1in,height=1.25in,clip,keepaspectratio]{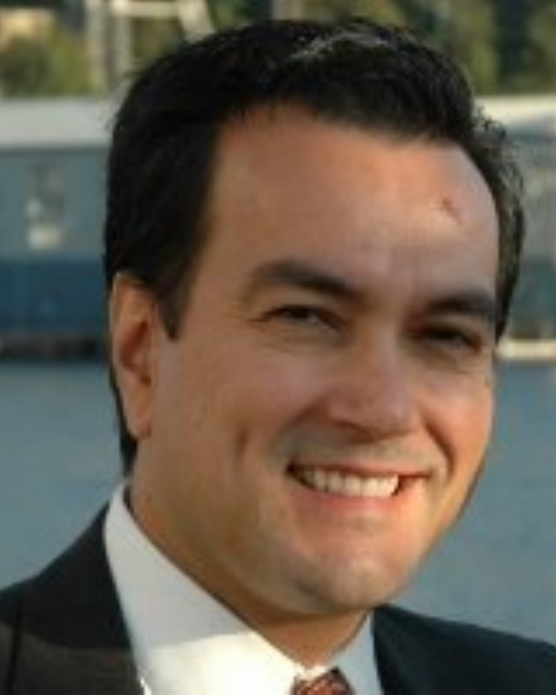}}]{Dinei Florencio} (M'96-SM'05-F'16) received the B.S. and M.S. degrees from the University of Bras\'ilia and the Ph.D. degree from Georgia Tech, all in electrical engineering. From 1996 to 1999, he was a member of the Research Staff, David Sarnoff Research Center. He has been a Researcher with Microsoft Research since 1999. His research has enhanced the lives of millions of people, through high impact technology transfers to many Microsoft products, including Internet Explorer, Skype, Live Messenger, Exchange Server, RoundTable, and the MSN toolbar. He has authored over 100 referred papers. He holds 61 granted U.S. patents. He was General Co-Chair of the MMSP'09, Hot3D'10 and '13, and WIFS'11, and Technical Co-Chair of the WIFS'10, ICME'11, MMSP'13, and GlobalSIP'18. Dr. Florencio was a member of the IEEE SPS Technical Directions Board from 2014 to 2015 and the Chair of the IEEE SPS Technical Committee on Multimedia Signal Processing.
\end{IEEEbiography}
\vspace{-9 mm}

\begin{IEEEbiography}[{\includegraphics[width=1in,height=1.25in,clip,keepaspectratio]{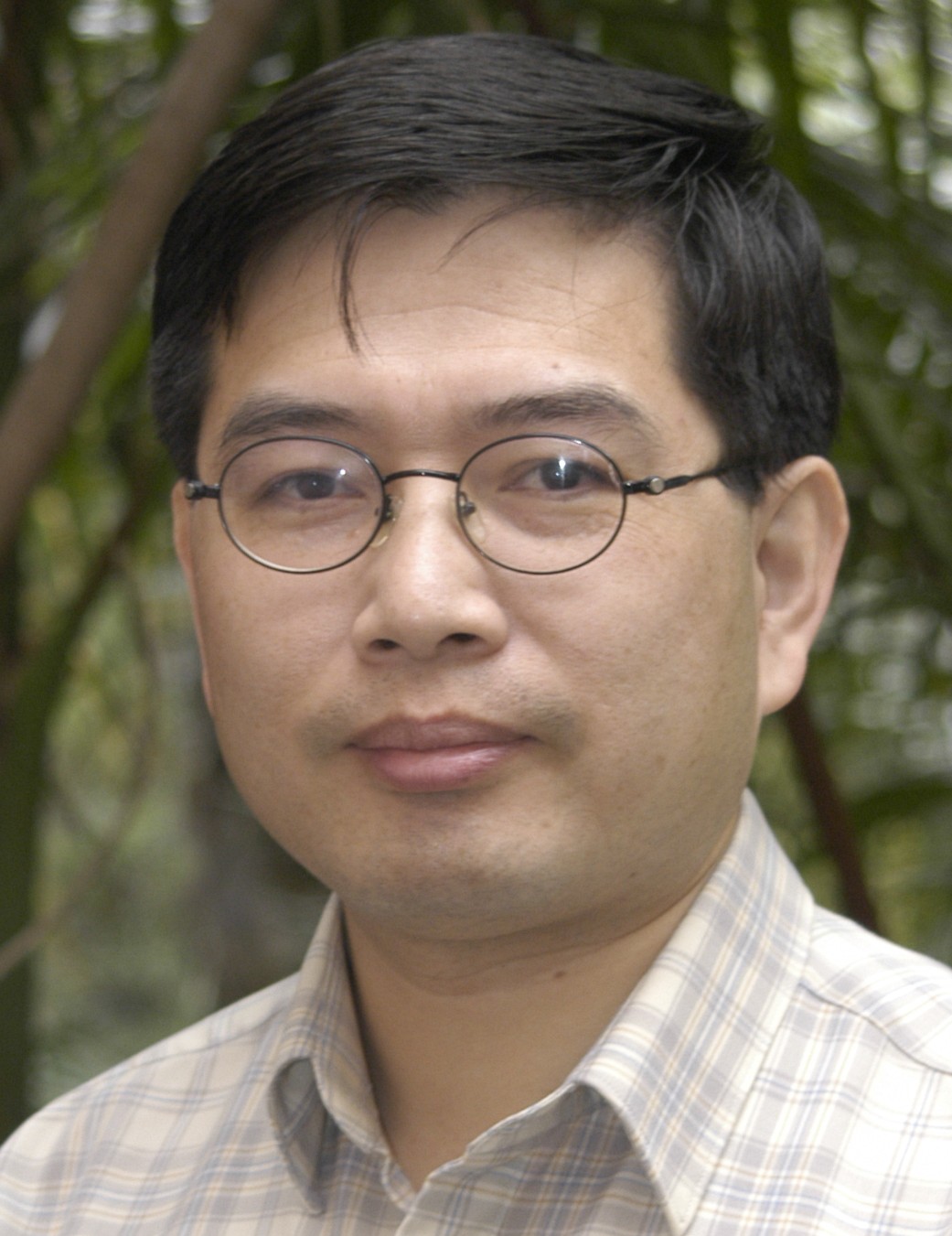}}]{Wanqing Li} received his Ph.D. in electronic engineering from The University of Western Australia. He was a Principal Researcher (98-03) at Motorola Research Lab and a visiting researcher (08, 10 and 13) at Microsoft Research US. He is currently an Associate Professor and Co-Director of Advanced Multimedia Research Lab (AMRL) of UOW, Australia. His research areas are 3D computer vision, 3D multimedia signal processing and medical image analysis. Dr. Li is a Senior Member of IEEE.
\end{IEEEbiography}
\vspace{-9 mm}

\begin{IEEEbiography}
[{\includegraphics[width=1in,height=1.25in,clip,keepaspectratio]{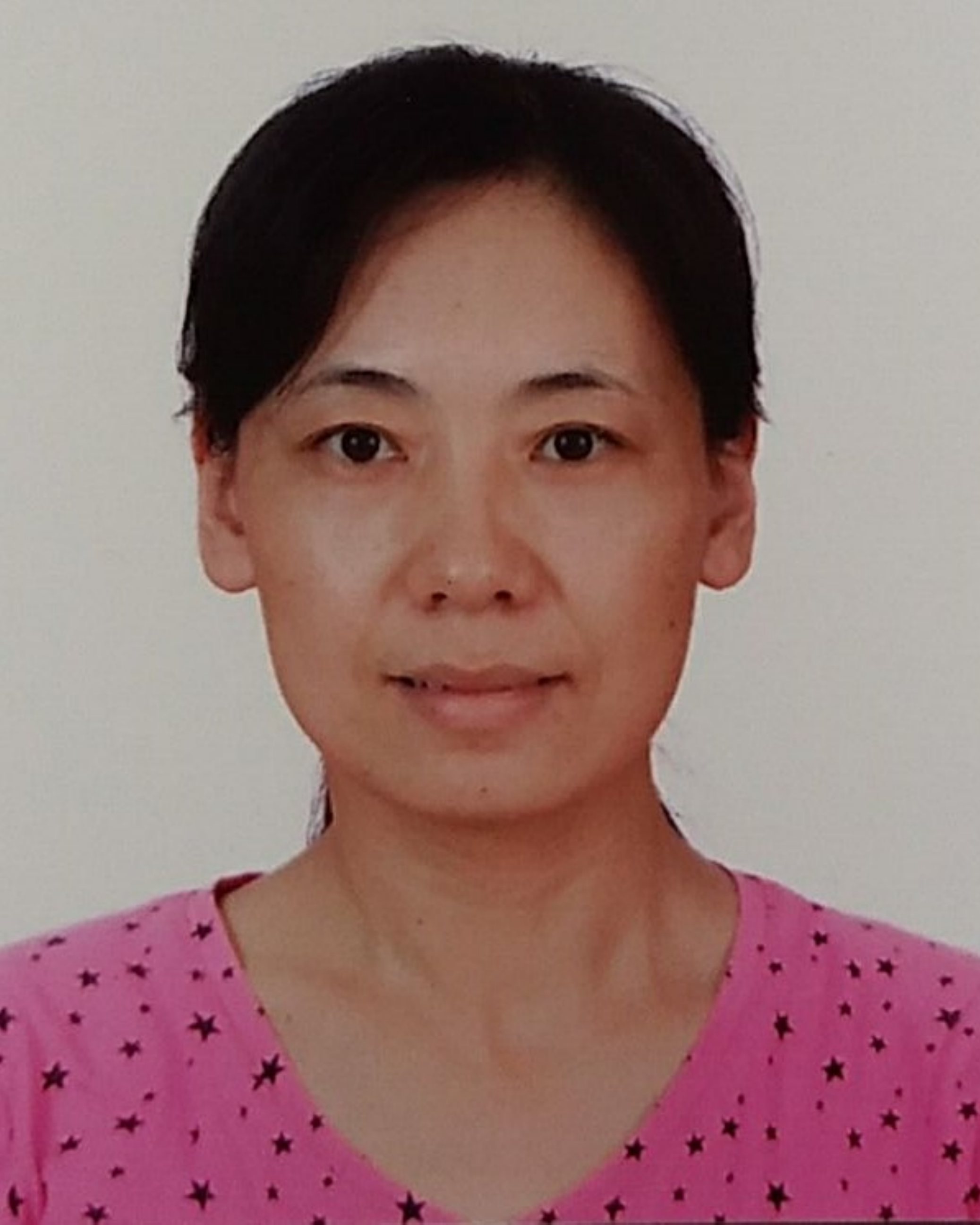}}]{Yaqin Zhao}
received her PhD from Nanjing University of Science and Technology in 2007. She is a professor with the College of Mechanical and Electronic Engineering, Nanjing Forestry University. She was a visiting fellow at the University of Wollongong in 2014. Her research interests include image processing and pattern recognition. 
\end{IEEEbiography}
\vspace{-9 mm}

\begin{IEEEbiography}[{\includegraphics[width=1in,height=1.25in,clip,keepaspectratio]{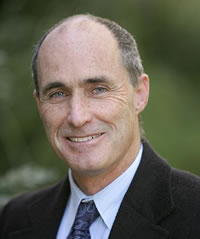}}]{Chris Cook}
is an electrical engineer with a BSc and BE from the University of Adelaide, and he received his PhD from the University of New South Wales in 1976. He is a Fellow of the Institute of Engineers Australia and a Chartered Engineer. He has worked for Marconi Avionics in the UK, for GEC Australia and was the founding Managing Director of the Automation and Engineering Applications Centre Ltd which designed and built industrial automation systems. He recently retired after 14 years as Executive Dean of Engineering and Information Sciences at the University of Wollongong but remains an active researcher in robotics and machine intelligence.  
\end{IEEEbiography}

\end{document}